# *DropSample*: A New Training Method to Enhance Deep Convolutional Neural Networks for Large-Scale Unconstrained Handwritten Chinese Character Recognition

Weixin Yang[a], Lianwen Jin[a,*], Dacheng Tao[b], Zecheng Xie[a], Ziyong Feng[a]

[a]College of Electronic and Information Engineering, South China University of Technology, Guangzhou, China

[b]Centre for Quantum Computation & Intelligent Systems, University of Technology, Sydney, Australia

**Abstract:** Inspired by the theory of Leitner's learning box from the field of psychology, we propose *DropSample*, a new method for training deep convolutional neural networks (DCNNs), and apply it to large-scale online handwritten Chinese character recognition (HCCR). According to the principle of *DropSample*, each training sample is associated with a quota function that is dynamically adjusted on the basis of the classification confidence given by the DCNN softmax output. After a learning iteration, samples with low confidence will have a higher probability of being selected as training data in the next iteration; in contrast, well-trained and well-recognized samples with very high confidence will have a lower probability of being involved in the next training iteration and can be gradually eliminated. As a result, the learning process becomes more efficient as it progresses. Furthermore, we investigate the use of domain-specific knowledge to enhance the performance of DCNN by adding a domain knowledge layer before the traditional CNN. By adopting *DropSample* together with different types of domain-specific knowledge, the accuracy of HCCR can be improved efficiently. Experiments on the CASIA-OLHDWB 1.0, CASIA-OLHWDB 1.1, and ICDAR 2013 online HCCR competition datasets yield outstanding recognition rates of 97.33%, 97.06%, and 97.51% respectively, all of which are significantly better than the previous best results reported in the literature.

**Keywords:** Convolutional neural network; deep learning; handwritten character recognition; domain-specific knowledge.

## 1. Introduction

A traditional isolated online handwritten Chinese character recognition (HCCR) method typically employs the following framework: (1) pre-processing of an input handwritten character (e.g., linear or non-linear normalization [2] and addition of imaginary strokes [3]-[7]), (2) feature extraction (e.g., 8-directional feature extraction [6] and discriminative directional feature extraction [8]), and (3) classification via machine learning methods (e.g., modified quadratic discriminant function (MQDF) [9][10], support vector machine [11], and hidden Markov model (HMM) [12]). In contrast, deep learning methods [13]–[17], which have attracted a considerable amount of research and industry attention in recent years, deviate from the above-mentioned framework by providing an alternative end-to-end solution to HCCR without any dedicated feature extraction or pre-processing technique, enabling potentially high performance. Owing to the availability of large-scale training data, new training technologies (e.g., Dropout [18], DropConnect [19], layer-wise pre-training [20]), and advanced computing hardware platforms (e.g., GPU [21]), the convolutional neural networks (CNNs) originally proposed by LeCun in the 1990s [22][23] have been extensively investigated in recent years. The traditional CNN has been extended with deeper architectures (e.g., [24][25]; we refer to this variant of CNN as Deep CNN (DCNN) in this paper), advanced training technologies, and effective learning algorithms (e.g., [18][40]) to address the various challenges posed by computer vision and pattern recognition problems. Consequently, significant breakthroughs have been achieved, such as image recognition [24]–[27], facial recognition [28][29], handwriting recognition [13][15][18][30], pose recognition [31], text detection, and natural scene image recognition [32]–[36]. Furthermore, DCNN with different structures has been successfully applied to the field of HCCR field [13]–[17], which poses a major challenge because it involves a large vocabulary (e.g., as many as 3755 classes for the GB2312-80 level-1 standard), many similar and confusable characters, and different writing styles with unconstrained cursive techniques [37].

The performance of current DCNNs is highly dependent on the greedy learning of model parameters via many iterations on the basis of a properly designed network architecture with abundant labeled training data. Most DCNN models treat all the training samples uniformly during the entire learning process. However, we have found that the error reduction rate during the learning process is initially high but decreases after a certain number of training iterations. This may be attributed to most of the samples being well recognized after a certain number of training iterations; thus, the error propagation for adapting the network parameters is low, while confusable samples, which are difficult to learn and account for a relatively low ratio of the training dataset, do not have a high likelihood of contributing to the learning process. Some previous DCNNs consider this phenomenon as a signal to manually reduce the learning rate, or as a criterion for early stopping [38], thereby neglecting the potential of the confusing samples that have thus far been insufficiently well-learnt.

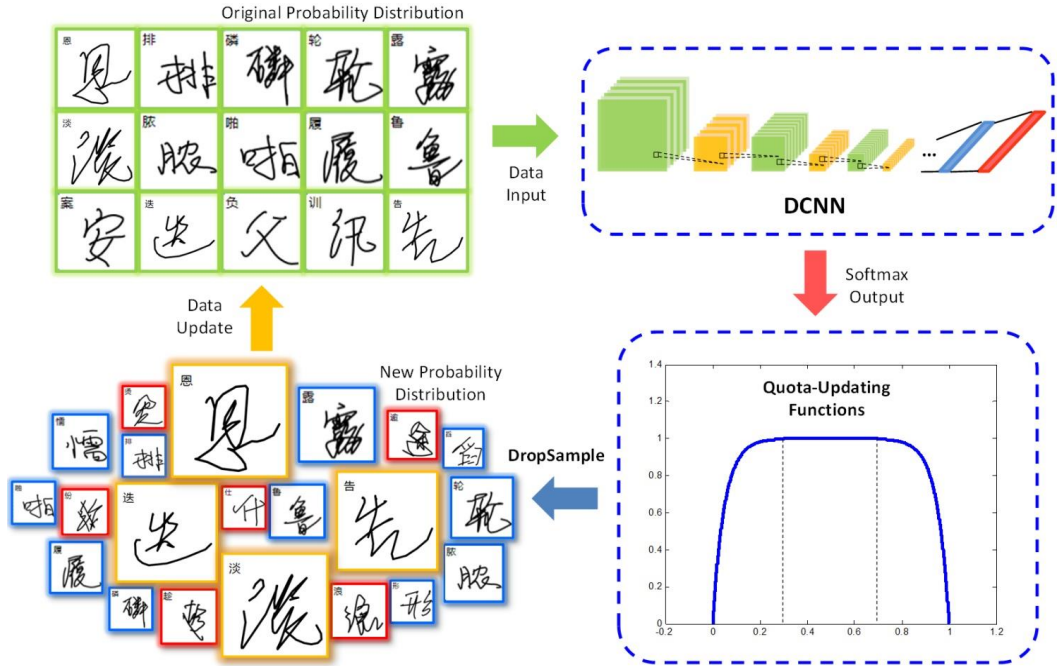

Figure 1 Illustration of *DropSample*. Initially, a uniform probability (also called quota) is selected for each training sample (upper left). The DCNN softmax outputs are used to separate the training set into three groups (blue for well-recognized samples, yellow for confusing samples, and red for heavily noisy or mislabeled samples). The selected probability distributions are then updated by the quota-updating functions.

Inspired by the theory of Leitner's learning box from the field of psychology [1], we propose a new training method, namely *DropSample*, to enhance the efficiency of the learning process of DCNN, and we employ it to solve the challenge of cursive online HCCR. Leitner's learning box is a simple implementation of the principle of spaced repetition for learning [39], which incorporates increasing intervals of time between subsequent reviews of previously learnt material in order to exploit the psychological spacing effect. Although the principle of spaced repetition is useful in many contexts, it requires a learner to acquire a large number of items and retain them in memory indefinitely. A direct application of Leitner's learning box theory is that material that is difficult to learn will appear more frequently and material that is easy to learn will appear less frequently, with difficulty defined according to the ease with which the user is able to produce a correct learning response. The *DropSample* training method proposed in this paper adopts a similar concept to design a learning algorithm for DCNN. To this end, each training sample is assigned to a box with a quota function that is dynamically adjusted according to the classification confidence given by the DCNN softmax output. After a learning iteration, samples with high confidence will be placed in a box with low appearance probability, whereas those with low confidence will be placed in another box with high appearance probability, thus they are more likely to appear for selection as the training data in the next iteration. A small amount of noisy data (e.g., mislabeled samples and outliers) always exists in the training dataset, and such data may be useful in the initial stages of training to avoid overfitting; however, they gradually prevent the network from achieving high prediction accuracy. They should therefore be placed in a box with low appearance probability, and thus can gradually be eliminated. A schematic of *DropSample* is shown in Figure 1.

To address the specific challenge of online HCCR, we propose the incorporation of domain-specific technologies in DCNN models to account for domain-specific information that may be useful but cannot be learnt by the DCCN. By employing the new training technology of *DropSample* together with various types of domain-specific knowledge, we find that recognition accuracy can be improved significantly. In this way, we obtain a single network that achieves test error rates of 3.07% and 3.45% on CASIA-OLHWDB 1.0 and CASIA-OLHWDB 1.1 [41], respectively, both of which are lower than the state-of-the-art results reported in previous studies [13][15][37]. Furthermore, the different types of domain-specific knowledge contribute to corresponding DCNN classifiers, which may be complementary and can therefore be integrated to achieve better accuracy. The ensemble ultimately reduces the test error rates to 2.67% on CASIA-OLHWDB 1.0, 2.94% on CASIA-OLHWDB 1.1, and 2.49% in the case of the ICDAR 2013 online HCCR competition dataset [16], which are significantly better than the best results reported in previous studies [13][15][16][37]; this confirms the effectiveness of the proposed *DropSample* training method.

The remainder of this paper is organized as follows. Section 2 introduces related studies reported in the literature. Section 3 presents the proposed DCNN architecture and the configurations employed. Section 4 provides a detailed description of the *DropSample* training method. Section 5 describes the domain-specific knowledge. Section 6 presents the experimental results and analysis. Lastly, Section 7 summarizes our findings and concludes the paper.

## 2. Related work

Over the past four decades, numerous studies have investigated HCCR [42], resulting in the development of many techniques such as non-linear normalization, data augmentation, directional feature extraction, and quadratic classifier modification. Non-linear normalization methods such as modified centroid-boundary alignment (MCBA) [43], line density projection interpolation (LDPI) [2], and line density projection fitting (LDPF) [43], as well as their pseudo-two-dimensional extensions [2][44], are based on line density equalization that can reduce within-class variation of character shape. Data augmentation techniques enhance insufficient training data by generating various handwriting styles via affine transformation [45], cosine functions [46], distorted generation [47], deformation transformation [48], and style consistent perturbation [57]. Two of the most popular feature extraction methods for representation are 8-directional feature extraction [6][10] and discriminative directional feature extraction [8][49]. By employing classifiers that use the modified quadratic discriminant function (MQDF) [9][10], or its variant such as discriminative learning quadratic discriminant function (DLQDF) [37], a traditional HCCR system can yield fairly good recognition performance.

In recent years, CNNs have produced outstanding results in the fields of machine learning and pattern recognition. The idea of CNN was first proposed by Fukushima [50] in 1980. It was formally developed by LeCun et al. [22][23] and improved by Simard et al. [51], Cireşan et al. [13][14], and others. GPU acceleration hardware [21] has facilitated the development of deep CNN (DCNN), which includes a deeper architecture with additional convolutional layers. DCNN offers several advantages. For example, it enables integrated training of feature extractors and classifiers to provide systematic optimization. In addition, it does not require pre-training and provides useful properties such as availability of raw data, effective feature extraction, and excellent generalization capability [52].

CNN has acquired a reputation for solving many computer vision problems in recent years, and its application to the field of HCCR has been shown to provide significantly better results than traditional methods [13]-[17]. The multi-column deep neural network (MCDNN) method proposed by Cireşan et al. [13] shows remarkable ability in many applications and attains near-human performance on handwritten datasets such as MNIST. In addition, it has provided promising results for HCCR. Graham proposed a variation of CNN called DeepCNet [15], which won first place at the ICDAR 2013 online HCCR competition [16]. By combining the path signature feature and employing a spatially sparse architecture, DeepCNet produced a best test error rate of 3.58% [15] on CASIA-OLHWDB 1.1, which is lower than that achieved by MCDNN (5.61%) [13] and DLQDF (5.15%) [37]. Furthermore, the use of the deep model for the recognition of digits [22] and various texts of different languages, such as English [53], Arabic [54], Tibetan [55], Hangul [52], and Devanagari [56], has also attracted considerable attention.

## 3. Architecture of sparse DCNN

DeepCNet [15], the variation of CNN proposed by Graham as mentioned above, exploits the spatial sparsity of the input layer to reduce computational complexity. The spatial sparsity means that an input online isolated character drawn as an $N \times N$ binary image shows sparse pen-color pixels compared to its background pixels. The computational complexity of the image and trajectory is $O(N^2)$ and $O(N)$, respectively; therefore, it is reasonable to set rules to reduce the calculation of the background pixels. On the other hand, because convolutional networks often encounter the issue of spatial padding, DeepCNet provides a solution that adds a substantial amount of padding to the input image at no additional calculation cost for the sparsity. Moreover, DeepCNet employs slow convolutional and max-pooling layers instead of fast layers in its architecture. Smaller filters and deeper layers are used to gradually reduce the size of the feature maps so that more spatial information can be retained to improve the generalization.

The DCNN employed in this study offers the advantages described above, and its structure is similar to that of DeepCNet employed in [15]. However, our DCNN model differs from DeepCNet [15] in the three aspects: (1) We apply a domain-specific knowledge layer immediately after the input layer to make use of the domain knowledge that is useful for HCCR but cannot be learnt by the CNN. Such domain knowledge includes 8-directional feature maps [6], character maps with both real and imaginary strokes [7], the deformation transformation of handwritten Chinese characters [48], and path signature feature maps [15][58]. (2) We add an extra fully connected (FC) layer before the final softmax layer of CNN. (3) Our model is much slimmer than the DeepCNet model used for HCCR in [15]; in other words, we use a smaller number of convolutional kernels in each layer, and thus, fewer parameters have to be learnt and stored.

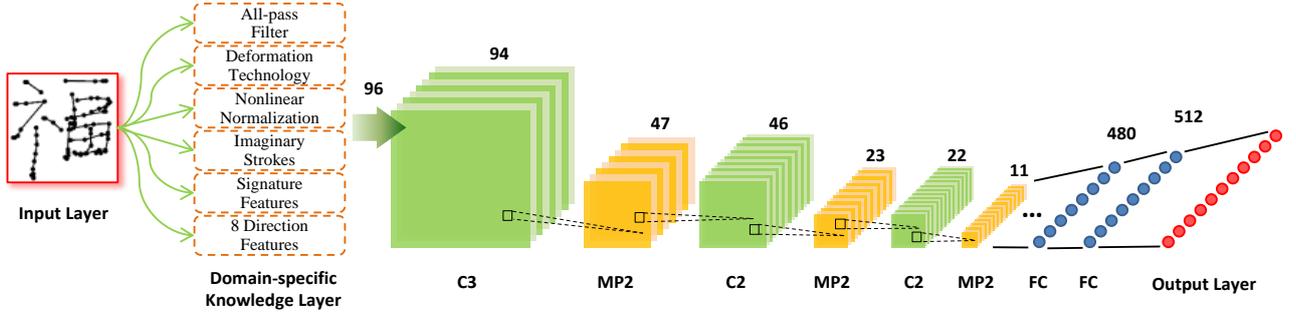

Figure 2 Illustration of the basic DCNN for HCCR.

As shown in Figure 2, the basic structure of our DCNN consists of five convolutional layers and two fully connected layers, while each of the convolutional layers is followed by a max-pooling layer. The size of the convolutional filter is 3×3 for the first layer and 2×2 for subsequent layers, with a stride of 1 pixel. Max-pooling is carried out over a 2×2 pixel window, with a stride of 2 pixels. Lastly, a stack of convolutional layers is followed by two FC layers containing 480 and 512 neurons, respectively. The number of convolutional filter kernels used is much smaller than is used in [15]; it is set as 80 for the first layer and increased in steps of 80 after each max-pooling. Consequently, the total number of parameters of our model is only 3.8 million, which is much smaller than the 5.9 million used by DeepCNet [15].

Rectified linear units (ReLUs) [59] are used as activation functions for neurons in all convolutional layers and FC layers. Compared to some saturating nonlinearities such as hyperbolic tangent [60] and sigmoid functions, non-saturating ReLUs can overcome the gradient diffusion problem and show better fitting ability when large-scale datasets such as handwritten Chinese characters are used for training [61]. In the output layer, the classification results are given by a $k$-way softmax layer, which outputs a probability distribution over the $k$ classes. Note that the softmax output can be treated as both an indicator of classification results and a confidence measurement of a given input sample [40]. The confidence measurement is discussed in detail in Section 4.

First, we render the input handwritten Chinese character image as a 48×48 bitmap image, then we embed the image in a 96×96 grid that is initialized to zero for spatial padding. The baseline architecture of our DCNN can be represented as

96×96Input-$M$×96×96-80C3-MP2-160C2-MP2-240C2-MP2-320C2-MP2-400C2-MP2-480FC-512FC-Output,

where $M$ denotes the number of input channels, which varies from 1 to 30 depending on the different types of domain knowledge used.

## 4. *DropSample*: A new training method for DCNN

### 4.1 Motivation

During the training process of DCNN for large-scale pattern recognition tasks, we usually encounter three problems.

First, although the problem of heavy computation in DCNN training can be alleviated using GPU-based massive parallel processing [51], DCNN training with an extremely large amount of training data for large-scale classification remains a time-consuming process. This is because the DCNN has to learn millions of parameters, and the convolution produces a large number of additional computations compared with traditional fully-connected shallow networks [22].

Second, the training error reduction during the training process is initially high but decreases after a certain number of training epochs (e.g., after 5-10 training epochs). This may be attributed to most of the samples being well recognized after a certain number of training epochs; thus, the error propagation for adapting the network parameters is low, while confusable samples, which are difficult to learn and account for a relatively low ratio of the training set, do not have a high likelihood of being selected for the training process. Some DCNNs treat this phenomenon as a signal for early stopping [38], thereby neglecting the potential of the confusable samples that have not thus far been sufficiently well-learnt and ignoring the proper processing of noisy samples (such as mislabeled samples or outliers).

Third, despite extensive efforts being devoted to training the dataset collection and cleaning, many databases include a significant number of mislabeled samples and heavily noisy samples. Figure 3 shows some examples of heavily noisy samples or mislabeled samples from the CASIA-OLHWB dataset [41]. Although the mislabeled or noisy samples constitute a minority of the overall dataset, they should not be tolerated. A primary reason is that after many training epochs, the error reduction rate gradually decreases and starts to oscillate. Thus, mislabeled or heavily noisy samples might be harmful because a strong error feedback produced by these samples will back-propagate from the output layer to previous layers and interfere with the entire network.

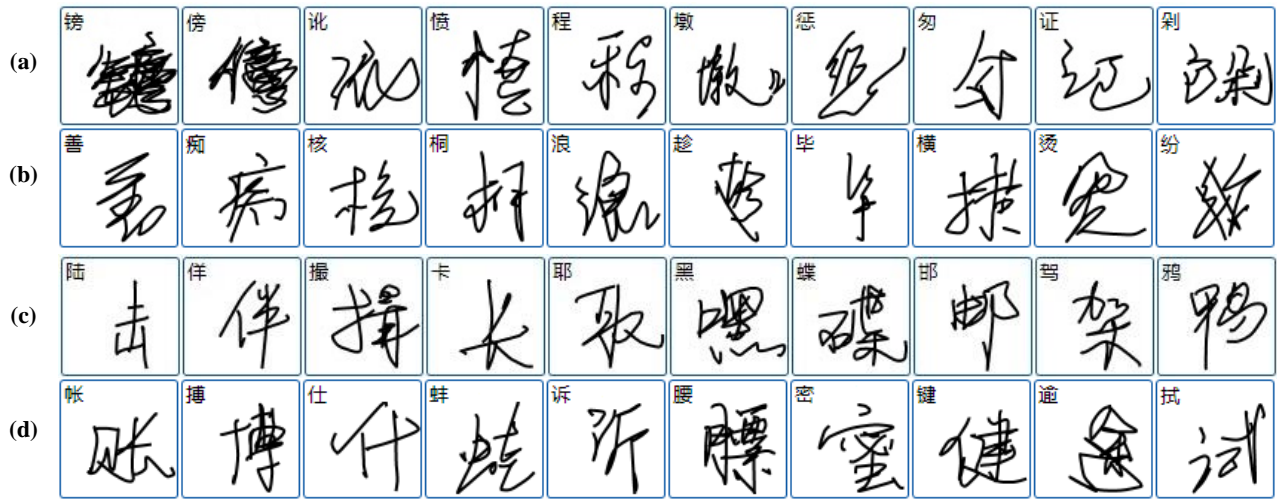

Figure 3 Examples of heavily noisy samples, mislabeled samples, and outliers from the CASIA-OLHWDB dataset. The given label of each example is shown in the upper left corner. (a) and (b) comprise heavily noisy samples or outliers, whereas (c) and (d) comprise mislabeled samples or outliers. Examples of (a) and (c) are taken from CASIA-OLHWDB 1.0, whereas (b) and (d) are from CASIA-OLHWDB 1.1.

To overcome the first two problems, Yuan et al. [62] developed a simple error samples reinforcement learning (ESRL) algorithm, which aims to make several interruptions during the training process and reconstruct the training set by randomly abandoning well-recognized samples and selecting several complex samples for data augmentation. Although this algorithm seems useful, it requires manual interruption, and the improvement is limited. In addition, the reconstruction of the sample distribution is delayed, and the abandoned well-recognized samples are no longer recalled for the training process. Moreover, the interference due to heavily noisy samples might be magnified during data augmentation.

Hermann Ebbinghaus, a German psychologist, discovered the forgetting curve, which describes the exponential loss of information that one has learned [63]. Inspired by his finding that the curve for meaningful material shows a slower decline than the curve for "nonsense" material, Sebastian Leitner proposed the Leitner system [1], which came to be widely used in the field of learning. In this system, a set of flashcards is sorted into groups according to how well the learner knows each flashcard. If the learner succeeds in recalling the solution written on a flashcard, the card is sent to the next group; otherwise, it is sent back to the previous group. For each successive group, the frequency with which the learner is required to revisit the cards decreases [1]. Inspired by this concept, we propose a new DCNN training solution to address the three problems mentioned above. The proposed training method is called *DropSample*; it sorts the training dataset into groups and uses the softmax output to dynamically and automatically adjust the selected quotas of samples in different groups. The output layer with $k$-way softmax as the activation function generates a probability distribution over the $n$ classes. According to a previous study, the softmax output of CNN can be regarded as a confidence measurement of the CNN classification output [40]. In the case of training using *DropSample*, samples with high confidence are placed in a box with low appearance probability and those with low confidence are discarded so that the remaining samples can be frequently reviewed in the network.

*4.2 Analysis*

Suppose that we have $m$ training samples $(x_i, y_i)$. The input vectors are $(n+1)$-dimensional. Batch gradient descent is used to train a DCNN with $k$-way softmax in the output layer. The hypotheses of softmax can be written as

$$h_{\theta_j}(x_i) = \frac{\exp(\theta_j^T x_i)}{\sum_{l=1}^{k} \exp(\theta_l^T x_i)} \quad (j=1,2,\ldots,k), \tag{1}$$

where $\theta_j$ denotes the weights and bias corresponding to the $j_{th}$ output. The loss function $J(\theta)$ with a regularization term can be written in the form of cross entropy as

$$J(\theta) = -\frac{1}{m}\left[\sum_{i=1}^{m}\sum_{j=1}^{k}1\{y_i = j\}\log\frac{\exp(\theta_j^T x_i)}{\sum_{l=1}^{k}\exp(\theta_l^T x_i)}\right] + \frac{\lambda}{2}\sum_{i=1}^{k}\sum_{j=0}^{n}\theta_{ij}^2. \tag{2}$$

Its partial derivative (known as the "error term") with respect to $\theta_j$ is given by

$$\nabla_{\theta_j}J(\theta) = -\frac{1}{m}\sum_{i=1}^{m}\left[x_i(1\{y_i = j\} - p(y_i = j | x_i;\theta))\right] + \lambda\theta_j, \tag{3}$$

where $p$ is the probability distribution of the softmax output, $1\{\cdot\}$ is the indicator function, and $\lambda$ is a penalty factor. Thus, one gradient descent iteration updates the parameters as

$$\theta_j := \theta_j - \alpha\nabla_{\theta_j}J(\theta) \quad (j=1,2,\ldots,k), \tag{4}$$

where $\alpha$ is the learning rate of the parameters.

According to (3) and (4), the error term can be split into three groups based on the softmax output. Let $p_i$ be the probability of the predicted class of the $i_{th}$ sample. Let $T_1$ and $T_2$ be the thresholds for $p_i$ that roughly separate $m$ training samples into three groups, $M_1$, $M_2$, and $M_3$, corresponding to the well-recognized group ($T_2 < p_i \leq 1$), confusing group ($T_1 \leq p_i \leq T_2$), and noisy group ($0 \leq p_i < T_1$), respectively. Thus, (3) can be rewritten as

$$\begin{aligned}\nabla_{\theta_j}J(\theta) &= E_1 + E_2 + E_3 + \lambda\theta_j \\ &= -\frac{1}{m}\Bigg\{\sum_{i_1 \in M_1}\left[x_{i_1}(1\{y_{i_1} = j\} - p(y_{i_1} = j | x_{i_1};\theta))\right] \\ &\quad + \sum_{i_2 \in M_2}\left[x_{i_2}(1\{y_{i_2} = j\} - p(y_{i_2} = j | x_{i_2};\theta))\right] \\ &\quad + \sum_{i_3 \in M_3}\left[x_{i_3}(1\{y_{i_3} = j\} - p(y_{i_3} = j | x_{i_3};\theta))\right]\Bigg\} + \lambda\theta_j,\end{aligned} \tag{5}$$

where $E_1$, $E_2$, and $E_3$ represent the sub-error terms corresponding to the three above-mentioned sample groups, respectively, which are analyzed as follows.

(1) Well-recognized group $M_1$: Given that the values of the probability distribution $p$ of a well-recognized sample are similar to those of the indicator function, the parameter $\theta_j$ obtains a small update from the sub-error term $E_1$ according to (4) and (5). Well-recognized samples are less useful for improving the network because they produce very low error feedback; therefore, it is reasonable to reduce their opportunities for selection as training data in the next training iteration.

(2) Confusing group $M_2$: In contrast, a confusing sample has a relatively decentralized softmax probability distribution. Neither the probability value of the predicted class nor the probability values of similar classes can be ignored, because both have an obvious effect on $E_2$. Moreover, in the feature space, confusing samples often appear near the decision boundary and exert an influence on boundary optimization. Therefore, the adaptive quotas of the confusing samples, which reflect their opportunity for selection as training data, should be increased to achieve fast, reinforced learning.

(3) Noisy group $M_3$: It is sometimes useful to acquire a large amount of feedback back-propagated from the sub-error terms. However, an exception arises in the case of noisy samples, which often produce a larger sub-error term $E_3$ than $E_1$ or $E_2$. These samples should be excluded from the training process, but not initially. This is because the network itself will find it difficult to identify heavily noisy or mislabeled samples during early training epochs. Moreover, the noisy samples can be regarded as noise for enhancing the regularization of the DCNN at the beginning of training [40].

## 4.3 Implementation of DropSample

Each sample in the training dataset is equally allocated an initial quota fixed to 1, and an updating function is then defined to change this quota according to the softmax output. The quota of a sample represents the probability of it being selected as a training sample.

We develop a quota-updating strategy as

$$q_i^t = q_i^{t-1} f(p_i^t), \tag{6}$$

where $f(\cdot)$ denotes the quota-updating function, $p_i^t$ is the softmax probability of the predicted class given by the $i_{th}$ sample at the $t_{th}$ updating iteration, and $q_i^t$ is the adaptive quota of the $i_{th}$ sample at the $t_{th}$ updating iteration. This update equation takes the previous updating quota into account, to prevent excessively fast quota adjustment. After a few epochs, a quota gradually absorbs information on the previous training results. Therefore, a well-recognized sample has a low quota resulting from excellent performance every time it is selected.

In practical application, the function $f(\cdot)$ can be manually fitted to certain requirements or tasks. Here, we present an exponential function and a multi-level hierarchical function. The exponential piecewise function $f_1(\cdot)$ is given by

$$f_1(p_i^t) = \begin{cases} 1 - \exp(-\gamma p_i^t) & 0 \leq p_i^t < T_1 \\ 1 - \exp(-\beta(1 - p_i^t)) & T_2 < p_i^t \leq 1, \\ 1/q_i^{t-1} & \text{otherwise} \end{cases} \tag{7}$$

where $\beta$ and $\gamma$ are the slope factors. A higher value of $\beta$ or $\gamma$ indicates a steeper shape of the function. In (7), the three equations apply to the three training groups corresponding to $M_1$, $M_3$, and $M_2$, respectively.

The multi-level hierarchical piecewise function $f_2(\cdot)$ is given by

$$f_2(p_i^t) = \begin{cases} a_{1h} & \{L_{1h} \leq p_i^t \leq U_{1h} \mid 0 \leq L_{1h} < U_{1h} < T_1, h = 1,2,\ldots l_1, i \in M_1\} \\ a_{2h} & \{L_{2h} \leq p_i^t \leq U_{2h} \mid T_1 \leq L_{2h} < U_{2h} \leq T_2, h = 1,2,\ldots l_2, i \in M_2\}, \\ a_{3h} & \{L_{3h} \leq p_i^t \leq U_{3h} \mid T_2 < L_{3h} < U_{3h} \leq 1, h = 1,2,\ldots l_3, i \in M_3\} \end{cases} \tag{8}$$

where $L_h$ and $U_h$ denote the lower and upper boundary, respectively, of the $h_{th}$ level of the hierarchical function. The parameters $a_{1h}$, $a_{2h}$, and $a_{3h}$ are the updating factors of the $h_{th}$ level of the three sample groups. As discussed in Section 3, these three equations apply to the corresponding sample groups. Because $M_2$ is the group of confusing samples, we fix the parameter $a_{2h}$ to $1/q_i^{t-1}$ to maintain a value of 1 for the quotas. Thus, (8) can be rewritten as

$$f_2(p_i^t) = \begin{cases} a_{1h} & \{L_{1h} \leq p_i^t \leq U_{1h} \mid 0 \leq L_{1h} < U_{1h} < T_1, h = 1,2,\ldots l_1, i \in M_1\} \\ a_{3h} & \{L_{3h} \leq p_i^t \leq U_{3h} \mid T_2 < L_{3h} < U_{3h} \leq 1, h = 1,2,\ldots l_3, i \in M_3\}. \\ 1/q_i^{t-1} & \text{otherwise} \end{cases} \tag{9}$$

In our implementation, the parameters for $f_1(\cdot)$ in (7) are determined empirically through experiments as follows: $\beta = 400$, $\gamma = 600$, $T_1 = 1/k$, and $T_2 = 0.99$. The parameters for $f_2(\cdot)$ in (9) are set as follows: $l_1 = l_3 = 3$, $a_{11} = a_{31} = 0.9$, $a_{12} = a_{32} = 0.5$, and $a_{13} = a_{33} = 0.1$. The boundaries are $L_{11} = 0$, $U_{11} = L_{12} = 1/4k$, $U_{12} = L_{13} = 1/2k$, $U_{13} = T_1 = 1/k$; $T_2 = L_{31} = 0.99$, $U_{31} = L_{32} = 0.999$, $U_{32} = L_{33} = 0.9999$, and $U_{33} = 1$.

There are several noteworthy points, as follows.

(1) The quotas are initialized to 1 so that we can obtain the equivalent training set size at each iteration, as shown in Figure 4. It can be seen that the curve initially declines rapidly, and only a third of the total samples remain for further learning after a certain number of iterations (700,000 mini-batch iterations in this case); this intuitively indicates during the training process that training with *DropSample* is increasingly efficient.

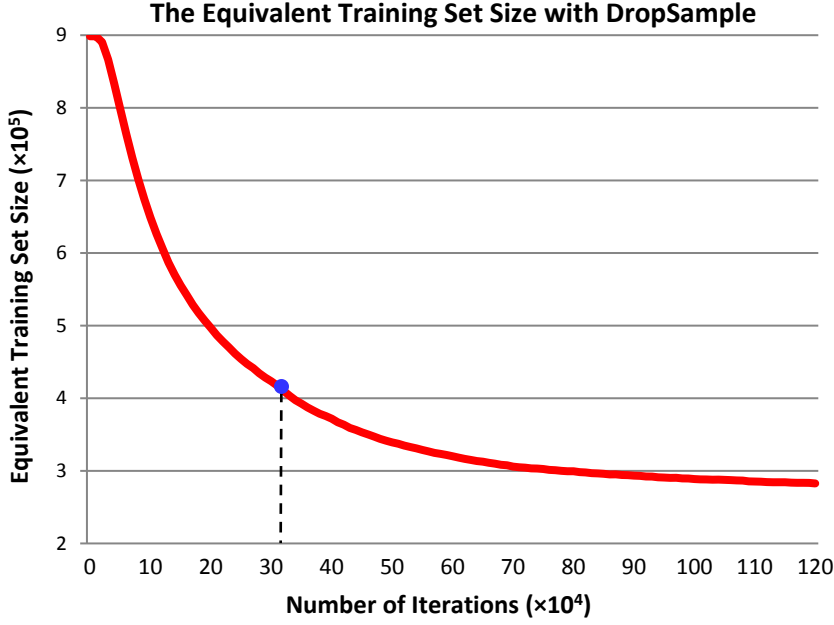

Figure 4 Illustration of the equivalent training set size when training with *DropSample*. The initial training set size is 898,524 on CASIA-OLHWDB 1.1. Heavily noisy samples are gradually eliminated after 300,000 mini-batch iterations in our implementation.

(2) *DropSample* does not actually drop samples from the training set; instead, it provides adaptive quotas for samples to dynamically adjust their opportunity for selection as training data. In other words, every sample has a chance to make a contribution.

(3) The threshold $T_1$, which defines the boundary of noisy samples, is set to $1/k$ because this value is equivalent to the random guess probability of $k$ classes. We consider a sample to be heavily noisy if the probability of its predicted class is lower than this threshold after a certain number of mini-batch training iterations (300,000 in our implementation, as shown in Figure 4, to ensure a fair judgment). Because we cannot strictly distinguish noisy samples, the random guess probability ensures that samples below this threshold are neither absolutely correct labeled samples nor characters similar to the true labeled sample. The threshold $T_2$, which defines the boundary of well-recognized samples, should be set to a high value close to 1.

(4) As the *DropSample* technique is independent of the network architecture, it shows high flexibility and can be extended to other deep models such as network in network [26], DBN [64], and stacked auto-encoder [65].

Finally, the proposed *DropSample* training algorithm is summarized in Algorithm 1.

---

**Algorithm 1** *DropSample* training algorithm

**Input:** training set $X = \{(x_i, y_i)\}, i = 1,...,m$ of $k$ classes.

**Output:** network parameters $\theta$.

**Initialization:** iteration $t \leftarrow 0$; learning rate $\alpha(t)$; quota parameters $q_i^0 \leftarrow 1, \forall i$;

quota-updating function $f_1$; $\beta = 400$, $\gamma = 600$, $T_1 = 1/k$, $T_2 = 0.99$.

1: **while** not converge **do**

2: $\quad P^t = (q_1^t / Z^t, ..., q_m^t / Z^t)^{\mathrm{T}}$ where $Z^t = \sum_{i=1}^{m} q_i^t$

3: $\quad t \leftarrow t+1$ sample a mini-batch from $X$ based on $P^t$

4: $\quad$ forward propagation: get the softmax output of the predicted class $p_i^t$

5: $\quad$ back propagation: calculate error term $\nabla_\theta J(\theta) = E_1 + E_2 + E_3 + \lambda\theta$

6: $\quad$ update network parameters $\theta = \theta - \alpha(t)\nabla_\theta J(\theta)$

7: $\quad$ calculate quota-updating function $f_1(p_i^t)$

---

|   |   |
|---|---|
| 8: | **if** $0 \leq p_i^t < T_1$, $f_1(p_i^t) = 1 - e^{-\gamma p_i^t}$ |
| 9: | **else if** $T_2 < p_i^t \leq 1$, $f_1(p_i^t) = 1 - e^{-\beta(1-p_i^t)}$ |
| 10: | **else** $f_1(p_i^t) = 1/q_i^{t-1}$ |
| 11: | update quota parameters $q_i^t \leftarrow q_i^{t-1} f_1(p_i^t)$ |
| 12: | **end while** |

## 5. Domain-specific knowledge layer of DCNN

Our DCNN architecture consists of a domain-specific knowledge layer that is used to process and embed useful domain knowledge, as shown in Figure 2. Previous studies have shown that the incorporation of domain-specific knowledge, such as path signature feature maps [15], is very useful for achieving highly successful online HCCR. This inspired us to adopt various types of domain knowledge in our DCNN models in this study, including deformation transformation [47][48], non-linear normalization [2], imaginary stroke maps [3], 8-directional feature maps [9], and path signature feature maps [58].

*5.1 Deformation transformation*

Deformation transformation (DT) for DCNN is used to provide shape variation and generate a substantial amount of online training data. It is very helpful for enhancing the performance of DCNN because a distribution that is considerably more representative can be roughly simulated by elaborately designing the transformation and applying it to the training samples in order to enhance the generalization capability of the network. In DeepCNet [15], the training set is extended by affine transformation, including global stretching, scaling, rotation, and translation, but only stroke jiggling is used for local deformation. To enrich the data with local diversity, two deformation methods are considered in this paper. The first is the distorted sample generation method [47], which can handle both shearing and local resizing. The second is deformation transformation [48], which provides adjustable parameters for shrinking or stretching parts of the character both locally and globally to generate rich handwriting styles.

*5.2 Non-linear normalization*

The non-linear normalization (NLN) method is based on line density equalization and allows shape correction, which has been shown to be a crucial pre-processing technique for traditional HCCR methods. In our implementation, we use an NLN method called line density projection interpolation (LDPI), which has been reported to show good performance in a previous study [2]. However, given that the max-pooling in DCNN can absorb positional shifts, the NLN step might be redundant or even result in information loss [52]. Moreover, the NLN step for data pre-processing may cause loss of diversity and reduce the generalization capability of the network. Therefore, our DCNN incorporates the NLN step before deformation transformation in the domain knowledge layer in order to retain the diversity of the training samples.

*5.3 Imaginary strokes technique*

Imaginary strokes (IS) [3] are those pen-moving trajectories in pen-up states that are not recorded in the original character sample, and experiments have shown that they facilitate effective HCCR [3]–[7]. An imaginary strokes map is defined as a character map that consists of all the straight lines from the end point of each pen-down stroke to the start point of the next pen-down stroke [6]. It is extremely difficult for a DCNN to learn such virtual information only from a 2D character map. Moreover, a common drawback of adding such virtual information to the prototype is that similarities may be introduced between readily distinguishable characters. Thus, a trade-off between imaginary strokes and original strokes is recommended, in which case the DCNN performs well. Therefore, our DCNN incorporates both real stroke maps and imaginary stroke maps in the domain knowledge layer. This approach is found to be useful even though the number of feature maps in the domain knowledge layer is doubled.

*5.4 Path signature feature maps*

Path signatures (Sign.), introduced by Chen [58] in the form of iterated integrals, can solve any linear differential equation and uniquely express a finite path. The path signature feature was first applied to the recognition of online handwritten characters by Graham [15]. In general, the first three (starting from the zeroth) iterated integrals of a signature correspond to the 2D bitmap (1 feature map), direction (2 feature maps), and curvature of the pen trajectory (4 feature maps), respectively. Considering the complexity of online HCCR, we use the first three orders of the signature (7 feature maps) as input features because the contribution of integrals beyond the third order is negligible. Consequently, the path signature provides 7 feature maps from the domain knowledge layer to the succeeding convolutional layers.

*5.5 8-directional feature maps*

Owing to their excellent ability to express stroke directions, 8-directional features (8Dir) [9] are widely used in HCCR. This technique extracts features from online trajectory points using their projections on 8 2D directions, and accordingly, 8 pattern images are generated and employed as feature maps in the domain knowledge layer. The number of directions can be flexibly simplified into 4 or detailed into 16 or more, but 8 directions are usually employed to strike a balance between complexity and precision.

# 6. Experiments

*6.1 Experimental Database*

We used the CASIA-OLHWDB 1.0 (DB 1.0) and CASIA-OLHWDB 1.1 (DB 1.1) [41] databases, which have been set up by the Institute of Automation, Chinese Academy of Sciences. DB 1.0 contains 3740 Chinese characters in the GB2312-80 standard level-1 set (GB1) obtained from 420 writers (336×3740 samples for training, 84×3740 for testing), whereas DB 1.1 contains 3755 classes in GB1 obtained from 300 writers (240×3755 samples for training, 60×3755 for testing).

The database for the ICDAR 2013 HCCR competition [16] comprises three datasets for isolated characters (CASIA-OLHWDB 1.0~1.2). All the data were annotated at the character level. The test dataset, which was published after the competition, was obtained from 60 writers who were not considered in DB 1.0~DB 1.2. Note that in our experiment, we only use a combination of DB 1.0 and DB 1.1 to evaluate the ICDAR 2013 HCCR competition dataset, because the classes of DB 1.2 are outliers of the 3755 classes in GB1.

*6.2 Network configurations of DCNN*

We use the same single DCNN structure for comparison in all experiments as that presented in Section 3, i.e., 96×96Input-$M$×96×96-80C3-MP2-160C2-MP2-240C2-MP2-320C2-MP2-400C2-MP2-480FC-512FC-3755Output, with only different $M$ to represent the different types of domain knowledge used. A random mix of affine transformations (scaling, rotations, and translations) [15] is adopted as the basic method for data argument during the training stage. The training mini-batch size is set as 96, and the dropout [66] rates for the last four weighting layers are experimentally determined and set as 0.05, 0.1, 0.3, and 0.2, respectively. We performed our experiments on a PC with Intel 3.4$Hz$ i7 CPU, 8G RAM and a GTX980 GPU.

*6.3 Investigation of different online features against baseline method for HCCR*

To ensure fair comparison, the baseline method (denoted by Bitmap) trains CNN by directly rendering an online handwritten Chinese character as an offline bitmap (1 feature map) as a training sample, without using any domain-specific knowledge, as in the case of [13]. Inspired by the promising result reported in [15], we compare the iterated-integrals signature feature in different orders (denoted by Sign.x) with the traditional widely used 8-directional feature [9]. By incorporating the popular online features with the Bitmap, the results are improved significantly, as shown in Figure 5. Among them, the second-order truncated signature feature (containing 7 feature maps and denoted by Sign.2) produces better results than the first-order truncated version (3 feature maps and denoted by Sign.1) and the 8-direction feature (8 feature maps and denoted by 8Dir), and shows almost the same performance as the third-order truncated signature feature (denoted by Sign.3) which contains many more feature maps (as many as 15). As most of the domain-specific knowledge used in this paper is extracted on the basis of online information from a handwritten sample, the domain knowledge in the following experiments is extracted and applied after incorporation with the Sign.2 online feature maps, for the sake of its higher performance and reduced storage.

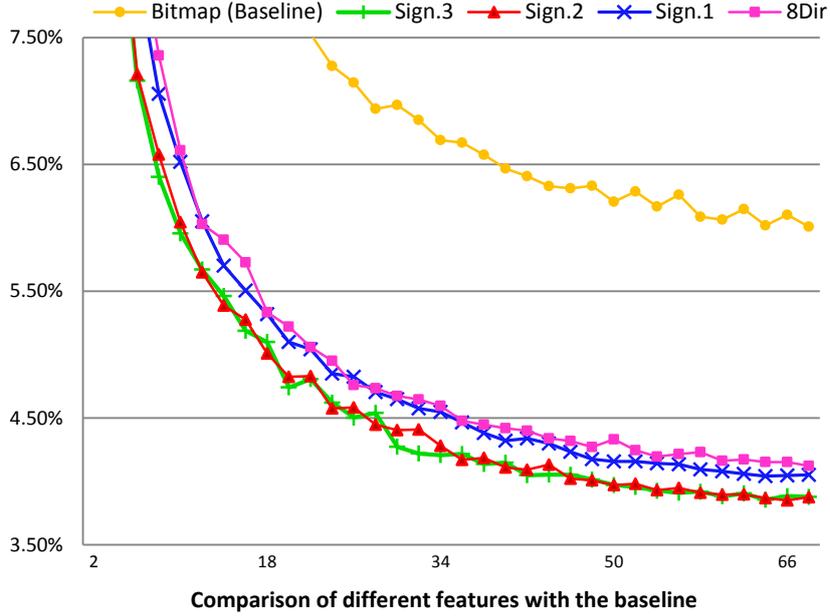

Figure 5 Comparison of different popular features. The x-axis represents the number ($\times 10^4$) of mini-batches for training and the y-axis denotes the test error rate.

### 6.4 Investigation of the effectiveness of domain-specific knowledge

We designed nine CNNs (denoted by A~I in Table 1) to intensively evaluate the performance achieved by the addition of different kinds of domain-specific knowledge to the baseline network. The experimental results for DB 1.1 are summarized in Table 1. From the results, the following interesting observations and conclusions can be made:

Table 1 Recognition rates (%) of different domain-specific methods on CASIA-OLHWDB 1.1

| Network | Domain-specific methods | Recognition rate (%) |
| --- | --- | --- |
| A | **Baseline**: Bitmap (no domain knowledge) | 93.99 |
| B | Bitmap + Sign.1 | 95.95 |
| C | Bitmap + Sign.2 | 96.12 |
| D | Bitmap + Sign.3 | 96.12 |
| E | Bitmap + Sign.2+ DT | 96.13 |
| F | Bitmap + Sign.2+ NLN | 95.81 |
| G | Bitmap + Sign.2+ 8Dir | 96.22 |
| H | Bitmap + Sign.2+ IS | 96.33 |
| I | **Fusion**: Bitmap + Sign.2 + DT + 8Dir + IS | **96.39** |

(1) The DCNNs with path signature features clearly enhance the Bitmap by adding online information. Network C using Sign.2 is better than Network B using Sign.1, and is also cost-effective compared to Network D using Sign.3 in terms of the number of feature maps. Similarly, all other networks with different domain knowledge outperform the baseline CNN by clear margins.

(2) Network E with DT shows a slightly better result than network C with Sign.2, while the improvement is more obvious in our preliminary small-scale experiments. Network F with LDPI [2] as the NLN method underperforms network C, although it produces better results than the other approaches in our preliminary experiments with small categories of training samples. As it was found that additional DT can somewhat extend the coverage of the possible handwriting styles, especially when data is insufficient, we keep the DT domain knowledge for further use by the ensemble model.

(4) Network G with 8-directional features produces better results than the baseline and Network C, indicating that the additional features offer extra statistical information on stroke direction to enhance the performance of the baseline CNN.

(5) Network H with imaginary strokes embedded in an online character shows an obvious improvement, indicating that the imaginary stroke technique is useful.

(5) By integrating all the domain knowledge (referred to as Fusion) except NLN to Network I, a significant improvement is achieved, which indicates that domain-specific knowledge is very useful for improving DCNN for HCCR.

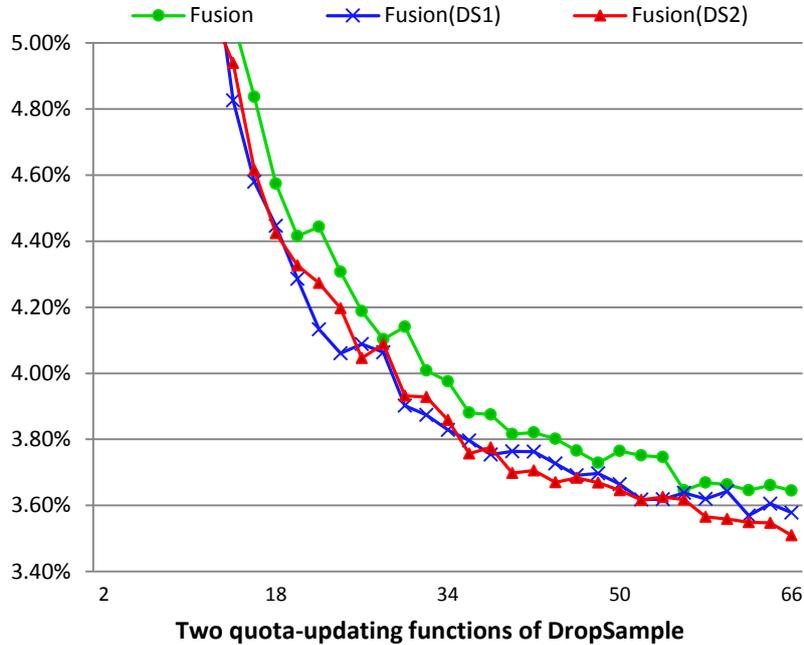

Figure 6 Performance of the two quota-updating functions implementing *DropSample* in terms of error rate. The x-axis represents the number ($\times 10^4$) of mini-batches for training.

*6.5 Investigation of DropSample training method*

First, we conducted experiments to compare the effects of the two quota-updating methods denoted in equations (7) and (9) (denoted by DS1 and DS2, respectively). Our experiments were deliberately evaluated over the fusion network which has already achieved high performance. The results are shown in Figure 6. It can be seen that even though the fusion network has already achieved promising results compared to the baseline, both *DropSample* methods further improve the performance. The multi-level hierarchy solution (DS2) shows slightly better results than the exponential decay solution (DS1).

We then applied the *DropSample* training method (DS2) to nine experiments with different combinations of domain knowledge. The results are shown in Figure 7 and Table 2.

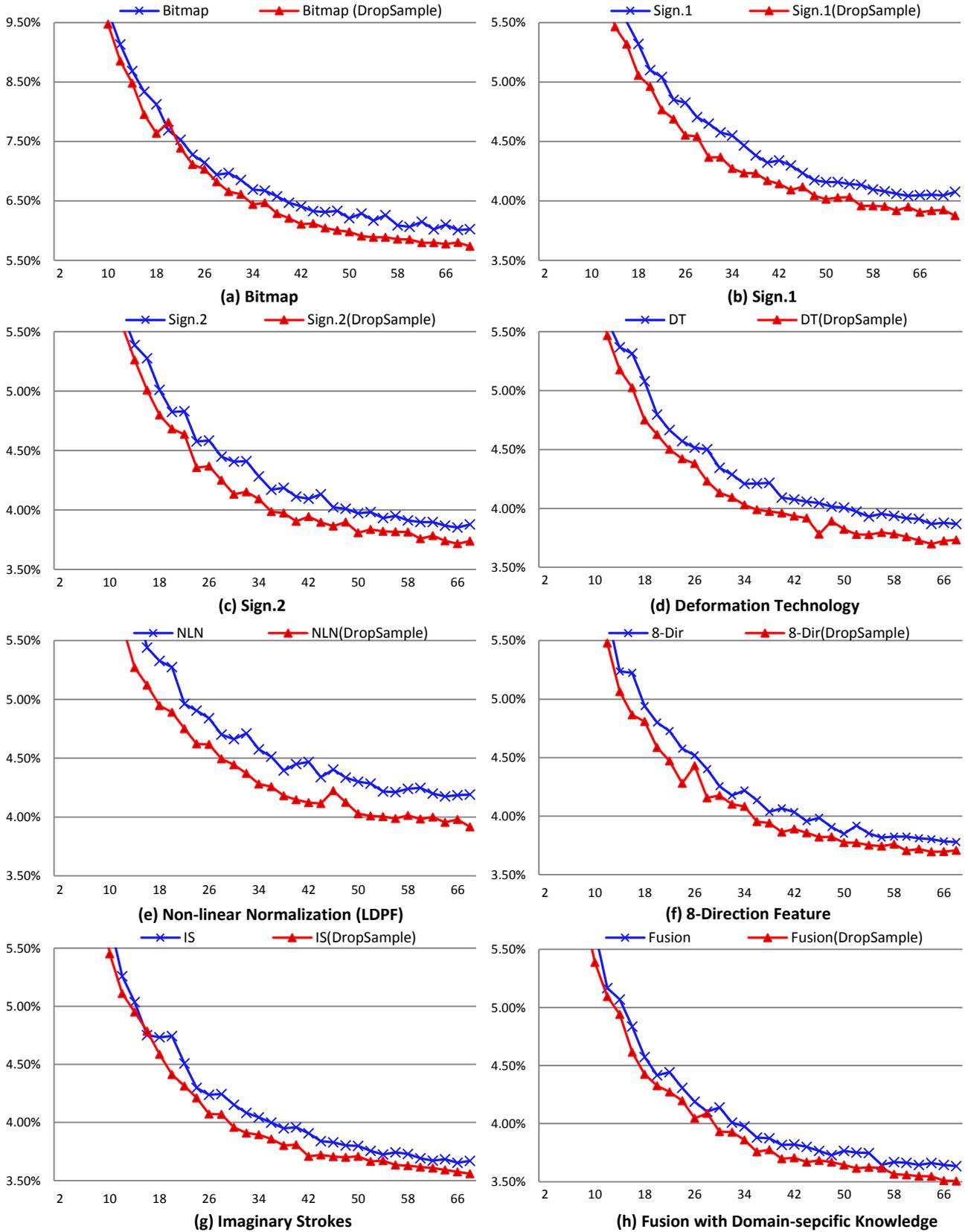

Figure 7 Performance of different types of domain knowledge used for training without or with *DropSample* method (denoted by DS) in terms of error rate. The x-axis represents the number ($\times 10^4$) of mini-batches for training.

Table 2 Recognition rates(%) for training with or without *DropSample* on CASIA-OLHWDB 1.1

| Network | Domain-specific methods | Without *DropSample* | With *DropSample* |
|---|---|---|---|
| A | **Baseline**: Bitmap | 93.99 | 94.20 |
| B | Bitmap + Sign.1 | 95.95 | 96.08 |
| C | Bitmap + Sign.2 | 96.12 | 96.26 |
| D | Bitmap + Sign.3 | 96.12 | 96.30 |
| E | Bitmap + Sign.2 + DT | 96.13 | 96.27 |
| F | Bitmap + Sign.2 + NLN | 95.81 | 96.08 |
| G | Bitmap + Sign.2 + 8Dir | 96.22 | 96.34 |
| H | Bitmap + Sign.2 + IS | 96.33 | 96.44 |
| I | **Fusion**: Bitmap + Sign.2 + DT + 8Dir + IS | 96.39 | 96.55 |

Table 2 shows that for all the experimental settings with different types of domain knowledge, DCNN trained with *DropSample* always outperforms DCNN trained without *DropSample*. From Figure 6, it can be seen that most of the samples at the beginning of training are fresh in the network, hence, they are treated as confusing samples and their quotas do not change over a period of several iterations. The performance curves with and without *DropSample* are therefore almost identical during the training period. However, the advantages of *DropSample* become apparent after the quotas of the well-recognized samples are gradually decreased, starting after some 150,000 iterations of mini-batch training.

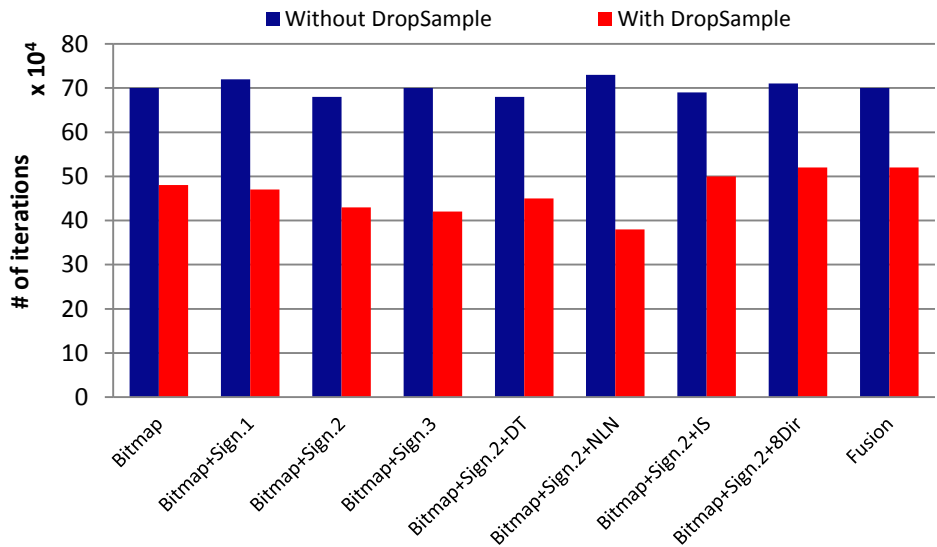

Figure 8 Comparison of the training efficiency with and without *DropSample*, in terms of the number of training interations involved for different networks.

We evaluate training efficiency with and without using *DropSample*. Figure 8 illustrates the number of training iterations required when both methods achieve the same recognition rates, which is the highest accuracy achieved by methods without using *DropSample*. It can be seen that the proposed *DropSample* training methods achieve the sample accuracies in much less time than the methods without *DropSample*, saving 25.71% ~ 47.95% training iterations. This clearly validates the high training efficiency of the *DropSample* method.

*6.6 Evaluation of ensemble of different types of domain knowledge embedded DCNNs for HCCR*

Given a set of DCNN models with different types of domain knowledge and training with and without *DropSample*, it is useful to combine the different types of domain knowledge to achieve better performance. Our ensembling method involves averaging the softmax output of the corresponding DCNN models. As shown in Table 3, we analyzed alternative ensembling methods, but they lead to inferior performance than simple averaging. The ensemble boosts the performance and achieves a high recognition rate of 97.33% on DB 1.0 and 97.06% on DB 1.1, with relative error reduction rates of 26% and 24%, respectively, compared to the baseline method. As shown in Table 4, our final result on DB 1.1 has a test error rate of 2.94%, which is significantly lower than that of the state-of-the-art DLQDF (5.15%) [37] and DeepCNet (3.58%) [15] approaches.

Additional experiments were conducted on the ICDAR 2013 HCCR competition dataset. As multiple classifiers vary among network architectures, configurations, and embedding techniques, their ensemble achieves a recognition rate of 97.51%, which is significantly better than that of the DeepCNet method (96.65%) with relative error reduction of 26%, and is also better than the winner of the ICDAR 2013 HCCR competition (97.39%).

Table 3 Recognition rates (%) of the single network and the ensemble on the three CASIA-OLHW datasets.

| Database | Single Network | | | Ensemble (Nets A~I) | | |
|---|---|---|---|---|---|---|
| | Baseline | Fusion | Training with *DropSample* | Max-pooling | Voting | Average |
| CASIA-OLHWDB 1.0 | 96.41 | 96.71 | **96.93** | 97.28 | 97.26 | **97.33** |
| CASIA-OLHWDB 1.1 | 96.12 | 96.39 | **96.55** | 96.97 | 96.99 | **97.06** |
| ICDAR 2013 competition DB | 96.65 | 96.93 | **97.23** | 97.40 | 97.43 | **97.51** |

Table 4 Comparison of different methods on the three CASIA-OLHW datasets.

| Database | Published State-of-the-art Performance (%) | | | | |
|---|---|---|---|---|---|
| | DLQDF [37] | MCDNN [13] | DeepCNet [15] | Our Single DCNN | Our Ensemble |
| CASIA-OLHWDB 1.0 | 95.28 | 94.39 | N/A | 96.93 | **97.33** |
| CASIA-OLHWDB 1.1 | 94.85 | N/A | 96.42 | 96.55 | **97.06** |
| ICDAR 2013 competition DB | N/A | N/A | 97.39* | 97.23 | **97.51** |

*\* The result of the winner of the ICDAR 2013 HCCR competition. Note that the number of parameters of DeepCNet (5.9 million) is much larger than that of our single DCNN (3.8 million).*

*6.7 Evaluation of large-scale online HCCR*

To show that the DCNN can be used to handle a very large-scale practical HCCR problem that involves a much greater number of classes (>3755), we extended the DCNN architecture to recognize as many as 10,081 classes of a mixture of handwritten Chinese characters, English letters, numerals, and symbols. The 10,081 classes of characters include 6763 simplified Chinese characters in the GB2312-80 standard, 5401 traditional Chinese characters in the Big5 standard, 52 English letters (26 upper-case and 26 lower-case letters), 10 standard numerals (0–9), 165 symbols, and 937 additional rarely used Chinese characters. Note that 3247 characters are common to both GB2312-80 and Big5, thus, there are 10,081 different classes of characters in all. The DCNN architecture we used is $96{\times}96\text{Input-}M{\times}96{\times}96\text{-}100\text{C3-MP2-200C2-MP2-}300\text{C2-MP2-400C2-MP2-500C2-MP2-600C2-1024FC-10081Output}$, which is similar to that presented in Section 3, but with more convolutional kernels. We use around 10 million data samples for training and 2 million data samples for validation, which were mainly obtained from the SCUT-COUCH dataset [67] and the CASIA-OLHWDB [41] dataset, together with some of our in-house datasets. It took us approximately two weeks to train and optimize this DCNN system[1]. Testing was conducted on another 827,685 samples randomly selected from seven datasets, as shown in Table 5; these samples were not included in the training or validation datasets. The results are summarized in Table 5. It can be seen that in spite of a very large number of classes (10,081), the DCNN model achieves a very promising recognition rate of 97.737% on average for a total of seven datasets. This result reflects the robust and excellent classification ability of the proposed DCNN with *DropSample* and domain knowledge enhancement.

Table 5 Performance of large-scale unconstrained handwritten character recognition on seven databases.

| Test Set | DB 1.0 [41] | DB 1.1 [41] | DB 1.2 [41] | SCUT-COUCH [67] | HKU [6] | In-house Dataset | 863 [68] | Total |
|---|---|---|---|---|---|---|---|---|
| Number of Samples | 143600 | 98235 | 79154 | 161166 | 120863 | 184089 | 40578 | 827685 |
| Recognition Rate(%) | 96.696 | 96.046 | 97.242 | 98.616 | 97.731 | 98.415 | 99.722 | 97.737 |

## 7. Conclusion

This paper proposed *DropSample*, a new training method for DCNN, through the efficient use of training samples. A DCCN that is trained with the *DropSample* technique focuses on confusing samples and selectively ignores well-recognized samples, while effectively avoiding interference due to noisy samples. We showed that most domain-knowledge-based processing methods in the field of HCCR can enhance DCNN via suitable representation and flexible incorporation. The recognition rates of our DCNN trained with *DropSample* significantly exceeded those of state-of-the-art methods on three publicly available datasets, namely, CASIA-OLHWDB 1.0, CASIA-OLHWDB 1.1, and the ICDAR 2013 HCCR competition dataset. Furthermore, the proposed DCNN was extended to very large-scale handwritten character recognition involving 10,081 classes of characters, with millions of training data, and a promising average recognition rate of 97.737% was achieved.

Although the *DropSample* method proposed herein has been designed for DCNN models to address the large scale HCCR problem, we expect that it will also serve as a general learning technique that can be extended to other pattern recognition tasks such as image recognition, as well as other deep learning models such as deep belief networks [64] and deep recurrent neural networks [69]. These potential applications merit further investigation.

---

[1] A real-time web demo of this DCNN online handwritten character recognition system is available at http://www.deephcr.net/.